\documentclass{article} 
\usepackage[preprint]{colm2025_conference}

\usepackage{microtype}
\usepackage{hyperref}
\usepackage{url}
\usepackage{booktabs}

\usepackage{lineno}
\usepackage{enumitem}
\usepackage{multirow}
\usepackage{graphicx}
\definecolor{mintgreen}{RGB}{235, 243, 237}
\definecolor{sagegreen}{RGB}{153, 188, 133}
\definecolor{lightyellow}{RGB}{254, 250, 224}
\definecolor{lightpink}{RGB}{247, 225, 215}
\usepackage[most]{tcolorbox}
\usepackage{colortbl}
\usepackage{wrapfig}
\newtcolorbox{promptbox}[2][]
{
  colframe = sagegreen,
  colback  = mintgreen,
  coltitle = white,  
  title    = {#2},
  #1,
}

\definecolor{darkblue}{rgb}{0, 0, 0.5}
\hypersetup{colorlinks=true, citecolor=darkblue, linkcolor=darkblue, urlcolor=darkblue}

\usepackage{algorithm}
\usepackage{algpseudocode}

\usepackage{subcaption}

\title{Context Selection and Rewriting for Video-based Educational Question Generation}

\author{
    Mengxia Yu, Bang Nguyen, Olivia Zino, Meng Jiang \\
    Department of Computer Science \\
    University of Notre Dame \\
    Notre Dame, IN 46556, USA \\
    \texttt{\{myu2,bnguyen5,ozino,mjiang2\}@nd.edu}
}

\begin{document}

\ifcolmsubmission
\linenumbers
\fi

\maketitle

\begin{abstract}
Educational question generation (EQG) is a crucial component of intelligent educational systems, significantly aiding self-assessment, active learning, and personalized education. While EQG systems have emerged, existing datasets typically rely on predefined, carefully edited texts, failing to represent real-world classroom content, including lecture speech with a set of complementary slides. To bridge this gap, we collect a dataset of educational questions based on lectures from real-world classrooms. On this realistic dataset, we find that current methods for EQG struggle with accurately generating questions from educational videos, particularly in aligning with specific timestamps and target answers. Common challenges include selecting informative contexts from extensive transcripts and ensuring generated questions meaningfully incorporate the target answer. To address the challenges, we introduce a novel framework utilizing large language models for dynamically selecting and rewriting contexts based on target timestamps and answers. First, our framework selects contexts from both lecture transcripts and video keyframes based on answer relevance and temporal proximity. Then, we integrate the contexts selected from both modalities and rewrite them into answer-containing knowledge statements, to enhance the logical connection between the contexts and the desired answer. This approach significantly improves the quality and relevance of the generated questions. Our dataset and code are released in \url{https://github.com/mengxiayu/COSER}.
\end{abstract}

\section{Introduction}
In-class quizzes are often used to engage students and assess their understanding of lecture content. The quizzes typically feature multiple-choice questions (MCQs) due to their objectivity, efficiency, and scalability. The growing prevalence of online learning has further increased demand for high-quality MCQs. However, manually producing such questions is a resource-intensive process that requires substantial time, expertise, and effort. Specialists must be involved to ensure that the questions are accurate, relevant, of appropriate difficulty, and cover the target spectrum of knowledge and skills \citep{mucciaccia2025automatic}. Automatic Question Generation (QG) systems offer promising solutions to alleviate this burden on educators. While existing QG techniques have shown progress in generating questions from structured textual content, extending these capabilities to lecture videos presents unique challenges. Unlike well-organized written materials, lecture videos contain unstructured, lengthy, and often noisy speech transcripts. Generating high-quality MCQs from such content requires precise alignment with target answers and specific timestamps in the lecture. Key challenges include: (1) selecting the most informative and relevant context from lengthy lecture transcripts and complementary slides;
(2) ensuring generated questions appropriately use the context and meaningfully incorporate the given answer.

Existing educational question generation (EQG) methods attempt to identify relevant contexts on which the generated questions are grounded \citep{ghanem-etal-2022-question}. Simplistic extraction heuristics result in either overly broad, irrelevant contexts, or overly narrow contexts that lack sufficient information, both of which negatively affect the quality of the generated questions \citep{noorbakhsh2025savaal}. Recent studies on long-context EQG \citep{wang-etal-2023-skillqg, ding-etal-2024-sgcm} performed context selection with supervised training, which relied on annotated training data and supervised finetuning.  Critically, existing EQG datasets \citep{chen2018learningq, gong2022khanq,hadifar2023eduqg,xu-etal-2022-fantastic} are mostly built from high-quality, predefined contexts (e.g. manually labeled segments from textbooks or carefully edited academic resources), or manually corrected transcripts of lecture videos. This setup does not realistically reflect typical educational settings. First, lecture speech in real-world classroom includes filler words, disfluencies, and auto-transcription errors. Second, the information conveyed in lecture speech is not as concise or structured as in textbooks. Third, the language used in the lecture could be informal, designed to elaborate on the formal knowledge provided in a set of slides.  
The mismatch between datasets based on idealized texts and real lecture speech poses significant challenges for the practical development and evaluation of EQG technology. To address this gap, we construct a dataset for video-based EQG that reflects a more realistic setting. Our dataset includes audio recordings of live lectures in college classrooms and the associated video recordings of their screens. We collect multiple-choice quiz questions by having educators watch the videos, pause them to create questions as desired, and record the associated timestamps. This dataset serves as a testbed for video-based EQG. We will release our dataset upon acceptance.

Our new dataset reveals a clear need for approaches that construct concise and relevant contexts from lecture videos for EQG. Large language models (LLMs) demonstrate the potential to address this challenge through their strong language modeling and long-context capabilities. That said, this paper introduces a novel LLM-based EQG framework specifically designed to:
(1) Dynamically select context segments given a specific timestamp and guided by a desired answer,
(2) Rewrite context segments to ensure clarity, conciseness, and explicitly incorporate the answer text,
and (3) Integrate multi-modal information from both textual data (e.g., audio-transcribed segments) and visual information (e.g., video frames of slides).
By effectively selecting and rewriting context, the framework improves the quality, specificity, and educational alignment of generated question stems.

In summary, this paper makes three key contributions:
\begin{itemize}
    \item A new dataset supporting in-class video-based EQG, which consists of lecture recordings and educator created timestamp-based MCQs.
    \item A novel framework named \textbf{COSER}, which integrates \textbf{Co}ntext \textbf{Se}lection and \textbf{R}ewriting explicitly tailored for answer and timestamp-aware EQG.
    \item A more reliable reference-based metric NLI score for question generation.
\end{itemize}

\section{Related Work}

\subsection{Educational Question Generation with LLM}
The rapid growth of large language models has shed light on automatic question generation to facilitate education. \citet{10.1007/978-3-031-11644-5_13} first explored EQG with ChatGPT under different basic settings, and found more in-context demonstrations for few-shot generation to be more effective. 
\citet{lee2024few} leveraged few-shot prompting with LLMs for EQG. MCQGen \citep{10577164} created answer-agnostic MCQs with chain-of-thought and self-refined strategies.
\citet{agrawal2024cyberq} constructed knowledge graphs from educational contexts, which were then used to design prompts for LLMs to generate questions for interactive learning. \citet{MAITY2025100370} investigated in-context learning strategies to generate questions better aligned with Bloom's revised taxonomy. However, none of the studies has explored improving noisy long context for video-based EQG. 

\subsection{Context Modeling for Question Generation}

Regardless of their long-context capabilities, LLMs are prone to be distracted by irrelevant context \citep{pan-etal-2024-contexts, wu2024how, 10.5555/3618408.3619699}. In answer-aware question generation, identifying answer-relevant contexts is important. Previous studies \citep{sun-etal-2018-answer,10.1145/3308558.3313737} predicted ``clue'' words based on their proximity to the answer, which performed well in generating questions from short contexts. \citet{ding-etal-2024-sgcm} trained a BART-based model to identify salient sentences as an auxiliary task for QG. \citet{li-zhang-2024-planning} leveraged an LLM to identify answer-containing sentences as key points when generating answer plans for QG. \citet{xia-etal-2023-improving} trained a model to plan content at fine-grained level of phrases and coarse-grained level of sentences, thus obtaining answer-relevant summaries of the context. The approach showed the effectiveness of incorporating answer spans with contexts in multi-hop QG. \citet{hadifar-etal-2023-diverse} ranked document segments by relevance and diversity. Savaal \citep{noorbakhsh2025savaal} was a multi-stage QG framework that retrieved and summarized information to improve conciseness and relevancy of contexts. Unlike previous approaches that require annotated data for fine-tuning, our study focuses on context selection and rewriting for zero-shot QG with LLMs.

\section{Dataset: AIRC}

\begin{table}[t]
    \small
    \centering
    \renewcommand{\arraystretch}{1.2}
    \setlength{\tabcolsep}{3pt}
    \begin{tabular}{ll|r|rrr|r|rrr}
    \toprule
     & &  \multicolumn{4}{c}{LLM-Frontier}  &\multicolumn{4}{|c}{DL-Intro} \\

    & & Mean   \tiny{$\pm$STD}&Min & Med. & Max  & Mean   \tiny{$\pm$STD}&Min & Med. &Max   \\
    \midrule
    Transcript & \# words & 3,447.6  \tiny{$\pm$2,003.6}&1,133& 3,069.0& 10,109& 10,075.7  \tiny{$\pm$2,187.5}&7,571& 9,699.0&13,145\\
    & \# seg.& 161.8  \tiny{$\pm$90.8}&40& 155.0& 508& 377.8  \tiny{$\pm$130.5}&207& 396.0& 540\\
 Keyframe& \# frames& 22.5  \tiny{$\pm$20.1}&1& 18.0& 69& 70.6 \tiny{$\pm$18.6}& 39& 73.5&94\\ 
    MCQ & \# choices & 4.0  \tiny{$\pm$0.0}&4& 4.0& 4& 3.9  \tiny{$\pm$0.4}&3& 4.0& 4\\
    \textbf{Question} & \# words & 12.9  \tiny{$\pm$4.9}&3& 4.8& 37& 14.4  \tiny{$\pm$8.9}&3& 12.0& 63\\
    Answer & \# words & 7.0 \tiny{$\pm$5.2}&1& 5.0& 40& 9.2  \tiny{$\pm$7.4}&1& 12.0& 29\\
    Distractor & \# words & 5.9  \tiny{$\pm$4.8}&1& 4.0& 40& 8.3  \tiny{$\pm$6.3}&1& 7.0& 31\\
    \bottomrule
    \end{tabular}
    \caption{Statistics of our AIRC dataset. \textbf{Question} stem is the target output of QG.}
    \label{tab:data_stats}
\end{table}

\subsection{Overview}
Existing EQG datasets such as LearningQ \citep{chen2018learningq}, KhanQ \citep{gong2022khanq}, and EduQG \citep{hadifar2023eduqg} primarily deal with short contexts. FairytaleQA \citep{xu-etal-2022-fantastic} includes long contexts from books, but not lecture transcripts.

We present \textbf{AIRC} (short for Artificial Intelligence in Real Classroom), a dataset for video-based EQG. The dataset consists of two college-level courses collected from real classrooms. One course, \textbf{LLM-Frontier}, is a graduate-level course about frontier research on Large Language Models. It consists of 27 research talk-style lectures, covering topics such as instruction tuning, pre-training, and reinforcement learning. The other course, \textbf{DL-Intro}, is an undergraduate-level course on deep learning. It consists of 8 one-hour lectures, covering various topics such as graph neural networks, computer vision, and language modeling. Descriptive statistics of our dataset are shown in Table \ref{tab:data_stats}. Compared to existing EQG datasets in Table~\ref{tab:dataset_comparison}, our dataset provides full recordings of live lectures in real classrooms, reflecting the practical challenges of EQG with long and noisy context.

\subsection{Data Collection}
We collected educational question data through a systematic process. First, we gathered authentic lecture recordings of the courses. These lectures were captured by the widely used platforms Zoom and YouTube, along with transcripts generated via their automatic captioning features. The video recordings primarily consisted of the instructors' screen-sharing sessions, typically displaying lecture slides. 

Next, annotators were instructed to review the lecture videos carefully. We recruited three volunteer annotators, one professor and two graduate students who have served as teaching assistants in related courses. They manually identified and documented timestamps associated with key instructional moments. For each timestamp, annotators created multiple-choice quiz questions designed to assess students' understanding of the associated context. To ensure the answer-aware QG task is optimizable and the collected questions can be used as references, we instructed the annotators to follow these guidelines: (1) ensure that the question is grounded on the lecture content, and (2) avoid meaningless answers, e.g., ``yes'',``no'', ``none of the above'', etc.

\begin{table}[t]
    \centering
    \small
    \renewcommand{\arraystretch}{1.2}
    \setlength{\tabcolsep}{2pt}
    \begin{tabular}{l|ccc|c|c|cc}
        \toprule
        \multirow{2}{*}{Dataset} & \multicolumn{3}{c|}{LearningQ} & \multirow{2}{*}{EduQG} & \multirow{2}{*}{FairytaleQA} & \multicolumn{2}{c}{AIRC (ours)} \\
        \cmidrule(lr){2-4} \cmidrule(lr){7-8}
        & TED-Ed & Khan-Video & Khan-Doc & & & LLM-Frontier & DL-Intro \\
        \midrule
        Source & \multicolumn{3}{c|}{e-Learning Platform} & e-Library &  e-Library & \multicolumn{2}{c}{Live Lectures} \\
        Materials & Videos & Videos & Textbooks & Textbooks & Storybooks & Videos & Videos \\
        Level & K12 & K12 & K12 & Undergrad & Children & Graduate & Undergrad \\
        Avg. Words & 847.6 & 1,370.8 & 1,306.6 & 12,641.5 & 2,313.4& 3,447.6 & 10,075.7 \\
        \bottomrule
    \end{tabular}
    \caption{Compared to existing EQG datasets with educator created questions, our dataset is under realistic settings for in-class quiz question generation.}
    \label{tab:dataset_comparison}
\end{table}

\subsection{Data Postprocessing}
For transcripts that do not have punctuation, we run a punctuation restoration model \citep{guhr-EtAl:2021:fullstop}. For visual information, we first crop the video frame to only include the lecturers' screen, then run keyframe detection to extract the slide images (See $\S$\ref{sec:keyframe_detection_algo} for details). Lastly, we associate the keyframes with transcript segments based on the timestamps. To obtain textual descriptions of slide images, we prompt an LLM, i.e. GPT-4o-mini \citep{achiam2023gpt}, to describe each extracted keyframe.

\section{Preliminary}

\subsection{Answer and Timestamp-Aware Educational Question Generation}\label{sec:problem_definition}
We formally define the task of answer and timestamp-aware question generation (ATEQG) as follows: Given a lecture video $V$, consisting of both audio-transcribed speech and visual keyframes of the lecturer's screen, the objective is to generate a meaningful and contextually relevant question conditioned upon a specified timestamp $t$ and a provided answer span $A$.

Specifically, the input lecture video comprises transcribed speech segments denoted by $S = [S_1, S_2, ..., S_N]$, where each segment $S_i \in S$ is a sequence of tokens $S_i = [s^{i,1},s^{i,2}, ..., s^{i,n_i}]$. Additionally, the visual modality of the video is represented by a set of visual keyframes $F=[F_1, F_2, ..., F_M]$, each corresponding to visual content captured from the lecturer's screen. Crucially, these keyframes and transcript segments are temporally aligned, such that each keyframe $F_j$ can be associated with one or multiple transcript segments $S_i$.

For any given timestamp $t\in[0,T]$, where $T$ denotes the total duration of the lecture video, and $t$ maps to a transcript segment $S_{i_t}$, where $i_t\in[1,N]$. The provided answer span $A$ is represented as a sequence of tokens $A=[a_1, a_2, ..., a_k]$.

The task then is to generate a natural-language question $Q=[q_1, q_2, ..., q_l]$ that is semantically coherent and contextually grounded in the aligned transcript segment $S_{i_t}$, its associated keyframes, and the specified span $A$. Mathematically, this task can be framed as modeling the following conditional probability:
$P(Q \mid \mathbf{S}, \mathbf{F}, t, A) 
= P([q_1, q_2, \dots, q_l] \mid C, A)$
, where the context $C$ is an informative representation derived from the transcript segments and keyframes relevant to the question.

\subsection{Context Construction for ATEQG}

We decompose the ATEQG task into two main stages: context construction and question generation. Given a question generation function $G_\phi$, which produces a question $Q$ conditioned on a context $C$ and answer span $A$, we have:
$Q = G_{\phi}(C, A)$.
In this formulation, the question generation parameters \(\phi\) remain fixed. Thus, context construction becomes the sole focus of optimization.
Formally, we aim to propose a context construction function $C_\theta$ that optimizes the expected conditional probability of the generated questions under the fixed QG function $G\phi$: 
\[
\theta^* = \arg\max_{\theta}\sum_{(\mathbf{S}, \mathbf{F}, t, A, Q)\in D}\log P_{\phi}(Q\mid C_\theta(\mathbf{S},\mathbf{F}, t, A), A)
\]
Here, the context construction function $C_\theta$  is responsible for constructing content from $\mathbf{S}$ and $\mathbf{F}$ to generate an optimal context $C$. 

\begin{figure}[t]
    \centering
    \includegraphics[width=\linewidth]{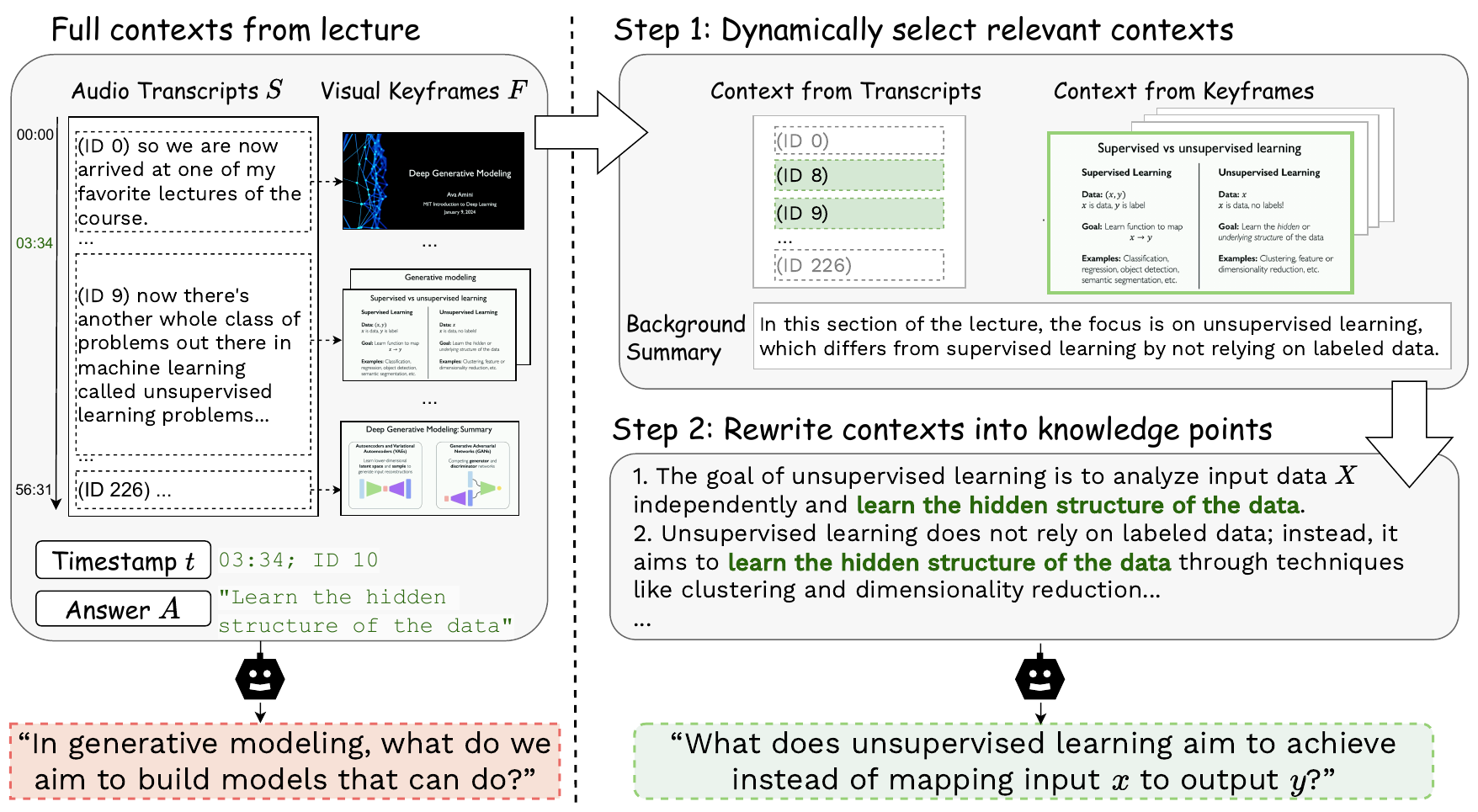}
    \caption{Our proposed framework COSER. Using all lecture content (left) as context results in generated questions that are too general and fail to incorporate keywords. COSER (right), which 
    (1) dynamically selects relevant contexts from both transcripts and keyframes, and (2) integrates and rewrites them into answer-containing knowledge points, yields more specific and relevant question.}
    \label{fig:method}
\end{figure}

\section{Method}
In this section, we propose COSER, an LLM-based framework for ATEQG. As illustrated in Fig.~\ref{fig:method}, COSER involves two main stages: (1) \textbf{Context selection} extracts transcript segments and visual keyframes conditioned on the given timestamp and answer span, and (2) \textbf{Context rewriting} revises the extracted context such that it is better suited for QG.

\subsection{Context Selection}

First, we prompt an LLM to select relevant context for creating a quiz question. Given a lecture represented by audio transcript segments $\mathbf{S}$ and visual keyframes $\mathbf{F}$, timestamp $t$, answer span $A$, the context selection process aims to identify a continuous subset of segments $S_{a:b} = [S_a, S_{a+1}, ... S_{b}]$, where $1\leq a \leq  b \leq N$, or keyframes $F_{u:v}=[F_u, F_{u+1}, ..., F_{v}]$, where $1\leq u \leq  v \leq M$. The extracted context should meet the following requirements: (1) \textit{integrity}: for transcript-based context, the output should be one or more unaltered segments directly extracted from the lecture transcripts; (2) \textit{relevancy}: it must be sufficient and concise, clearly providing all information required for creating the quiz question; (3) \textit{temporal proximity}: the selected segments might be near the given timestamp.

\subsection{Context Rewriting}
Next, with the extracted context, we ask the model to rewrite the context to incorporate the desired answer. Specifically, we ask the model to rewrite the context into several concise knowledge points or statements that contain the answer span in them. 

Formally, it can be expressed as:
$Rewrite_\theta\left(S_{p:q},F_{u:v}, t, A\right) = {K_1, K_2, ..., K_r}$
. Each knowledge statement $K_i$ is a sequence of tokens $K_i=[k_{i,1}, k_{i,2}],...,k_{i,m_i}$ that satisfies three requirements: (1) \textit{explicitness}: each statement $K_i$ explicitly includes the provided answer span $A: A \subseteq K_i, \forall K_i \in \{K_1, K_2, ...,K_r\}$; (2) \textit{atomity}: all statements should contain atomic knowledge and eliminate ambiguous references and indirect speech; (3) \textit{multi-granularity}: the set of statements should cover knowledge of different levels of granularity, ranging from conceptual or high-level knowledge to specific details.

The rewriting stage plays a crucial role in bridging the gap between raw lecture transcripts and question generation by converting implicit relationships into explicit statements and integrating the desired answers while maintaining natural language flow.

\subsection{Multi-Modal Integration}
To integrate information from both modalities, we first extract context from audio transcripts and visual keyframes separately. Then, we instruct the model to rewrite the transcript-based context into knowledge points, with context from keyframes as complementary information.

\section{Experiments}

\subsection{Evaluation}

\paragraph{NLI Score}

Recent QG studies \citep{luo-etal-2024-chain, ding-etal-2024-sgcm, wang-etal-2023-skillqg} adopt traditional natural language generation metrics such as BLEU, ROUGE, and BERTScore. However, these metrics often emphasize surface-level similarity and lexical overlap, which may not accurately reflect semantic fidelity in QG tasks \citep{mohammadshahi-etal-2023-rquge}. We aim to identify a metric that can reflect semantic fidelity of candidate question. To this end, we experiment on two benchmarks on paraphrasing questions: Minimal Edited Questions (MEQ) \citep{10.1162/tacl_a_00591} and Quora Question Pair (QQP) \footnote{https://www.kaggle.com/c/quora-question-pairs}. These benchmarks consist of positive pairs of questions that are paraphrases of each other, and negative pairs of questions that are not. We investigated the ability of QG evaluation metrics in accurately distinguishing the positive pairs of questions from negative ones. Specifically, we experimented with the following off-the-shelf scoring metrics: (1) Natural Language Inference (NLI) models, including three state-of-the-art DeBERTa-based models; (2) LLM zero-shot and few-shot; (3) Parascore \citep{shen-etal-2022-evaluation}; (4) ROUGE \citep{lin2004rouge}; (5) BLEU \citep{10.3115/1073083.1073135}; (6) BERTSsore \citep{zhang2020BERTScore}. Results presented in Table~\ref{tab:better_eval_metric} shows that NLI models outperform other metrics. NLI score evaluates the logical entailment between the candidate and the reference, thereby capturing deeper semantic relationships. We select the best-performing NLI model \footnote{https://huggingface.co/sileod/deberta-v3-large-tasksource-nli} as our reference-based metric, termed \textbf{NLI score}.

We also report a traditional reference-based metric \textbf{ROUGE-L}, and a reference-free metric \textbf{RQUGE} \citep{mohammadshahi-etal-2023-rquge}. RQUGE scores a question candidate from 1 to 5, based on its answerability given the context and target answer.
We allow LLM to generate up to 5 questions, and calculate NLI@5, RQUGE@5, and RougeL@5. We evaluate QG under zero-shot setting, using all data from LLM-Frontier and DL-Intro for testing.

\begin{table}[t]
\centering
\small
\begin{tabular}{l|rrr|rrr|rrr}
\toprule
 & \multicolumn{3}{c|}{GPT-4o-mini} & \multicolumn{3}{c|}{Llama-3.3-70B} & \multicolumn{3}{c}{Qwen-2.5-72B}\\
 & NLI & R-L  & RQG & NLI & R-L   &RQG& NLI & R-L   &RQG\\
 \midrule
\multicolumn{10}{l}{\textbf{Transcripts Only}}\\
All & 31.25 & \textbf{28.71}  & 3.18 & 24.05& 26.84 &2.94& 23.01& 27.30&2.61\\
Rule-Best & 28.75 & 22.90  & 2.97 & 32.21& 26.31& 3.03 & 29.77& 27.32& 2.88\\

Direct & 32.42& 25.70& 3.68 & 33.22& 27.46&3.85 & 32.81& 27.32&3.63\\
\rowcolor{mintgreen}
+ Rewrite & 34.80& 25.10& \textbf{4.86} & 33.44& 27.29&4.98 & 35.66& 28.75&4.88\\
CoT & 31.86 & 24.83  & 4.18 & 35.45& 29.31&4.31 & 33.36& 30.19&4.05\\
\rowcolor{mintgreen}
+ Rewrite & 34.42 & 24.80  & 4.85 & 35.33& 27.66& \textbf{4.99} & \textbf{37.76}& 29.43&\textbf{4.86}\\
\midrule
\multicolumn{10}{l}{\textbf{Keyframes Only}} \\
All & 21.58& 20.84& 2.52 & 15.48& 18.30& 2.29 & 15.46& 18.75&2.06\\
Rule-Best & 26.68 & 24.85  & 2.90 & 26.12& 25.46& 2.78 & 24.62& 23.50& 2.48\\
Direct & 28.22 & 23.27  & 2.99 & 28.23& 22.98& 3.09 & 28.10& 24.19&3.02\\
\rowcolor{mintgreen}
+ Rewrite & 29.07 & 23.90  & 2.82 & 29.65& 25.46& 4.83 & 31.12& 25.22&4.85\\
CoT & 26.35 & 24.17  & 2.96 & 28.08& 24.74& 3.17 & 26.36& 23.45&2.95\\
\rowcolor{mintgreen}
+ Rewrite & 30.84 & 25.13  & 3.13 & 31.28& 26.98& 4.85 & 30.32& 25.05&4.84\\
\midrule
 \multicolumn{10}{l}{\textbf{Multi-Modal}}\\
 CombineMM& 33.41& 27.12& 4.06 & 36.07& \textbf{31.04}&4.31 & 35.78& \textbf{30.83}&3.97\\
 \rowcolor{mintgreen}
 + Rewrite & \textbf{34.49}& 27.17& 4.80 & \textbf{36.19}& 28.10& 4.96 & 35.57& 30.00&4.80\\
 \bottomrule
  \multicolumn{10}{l}{\scriptsize{*Notes: R-L denotes Rouge-L F1; RQG denotes RQUGE. Higher means better for all metrics.}}
\end{tabular}
\caption{Results on LLM-Frontier.}
\label{tab:results_llm}
\end{table}

\subsection{Experimental Settings}
\paragraph{Context Settings}
We compare our method against the following context settings:
(1) \textbf{All}. Use the entire transcript or the complete set of keyframe descriptions as contexts, presented as an ordered list. Each segment is appended to an ordered ID. In the instruction, we also provide the segment ID indicating the timestamp.
(2) \textbf{Rule-$k$} .
Use the given timestamp to locate relevant content, then apply a fixed-length context window of size \(k\), where \(k\) denotes the number of transcript segments or keyframes. We experiment with \(k \in \{1, 3, 5, 7, 9, 11\}\). and report the best-performing setting as Rule-Best.
For our LLM-based context selection, we explore two strategies:
(3) \textbf{Direct} 
Provide the model with straightforward instructions for selecting relevant context.
(4) \textbf{Chain-of-thought (CoT)} 
After providing instructions, encourage LLM to explicitly output its reasoning process \citep{wei2022chain}. Specifically, we manually define the reasoning process as first listing all relevant segments, then refining its selection on a sentence-by-sentence basis. We also explore a basic combination strategy (\textbf{CombineMM}) where we concatenate segments selected by CoT from each modality.

\paragraph{Question Generation and Base Models}
We use a consistent prompt for QG for all context settings, which has been carefully engineered and optimized under Rule-Best settings. We allow the LLM to generate up to 5 questions.
We use GPT-4o-mini \cite{achiam2023gpt}, Llama-3.3-70B-Instruct-Turbo \citep{grattafiori2024llama}, Qwen-2.5-72B-Instruct-Turbo \citep{yang2024qwen2} as our base LLMs. We use the same LLM for context selection, rewriting, and question generation. See prompt templates in $\S$\ref{sec:qg_prompt}.

\section{Results and Analysis}
\begin{table}[t]
\centering
\small
\begin{tabular}{l|rrr|rrr|rrr}
\toprule
 & \multicolumn{3}{c|}{GPT-4o-mini} & \multicolumn{3}{c|}{Llama-3.3-70B} & \multicolumn{3}{c}{Qwen-2.5-72B}\\
 & NLI & R-L  & RQG & NLI & R-L   &RQG & NLI & R-L   &RQG \\
 \midrule
 \multicolumn{10}{l}{\textbf{Transcripts Only}}\\
All & 34.64& 29.86& 3.05 & 28.49& \textbf{28.69}& 2.91 & 25.45& 26.41&2.63\\
Rule-Best & 33.99& 28.35 & 2.97 & 38.32& 28.76& 3.25 & 37.17& 26.59& 3.14\\

Direct & 31.48& 24.04& 3.41 & 35.54& 26.21& 3.57 & 36.11& 25.93&3.40\\
\rowcolor{mintgreen}
+ Rewrite & 35.19& 26.53& \textbf{4.76} & 38.82& 26.69& \textbf{4.95}& 36.34& 27.71&4.79\\
CoT & 32.78& 24.66& 3.94 & \textbf{39.34}& 27.07& 3.87& 33.36& 30.19&3.96\\
\rowcolor{mintgreen}
+ Rewrite & 38.36& 27.36& 4.70 & 38.98& 27.60&4.94 & 37.33& 29.14&4.80\\
\midrule
\multicolumn{10}{l}{\textbf{Keyframes Only}} \\
All & 28.34& 23.76& 2.89  & 20.89& 18.22& 2.66 & 25.04& 22.02&2.65\\
Rule-Best & 31.23& 24.74& 2.47 & 24.23& 24.23& 2.73 & 26.40& 22.52& 2.53\\
Direct & 29.08& 24.94& 2.86 & 36.64& 26.04& 3.31 & 36.11& 25.93&3.06\\
\rowcolor{mintgreen}
+ Rewrite & 32.94& 22.05& 4.60 & 35.05& 24.67& 4.93 & 37.19& 26.60& \textbf{4.80}\\
CoT & 32.38& 25.27& 2.96  & 35.29& 25.10& 3.32 & 32.10& 24.38&3.07\\
\rowcolor{mintgreen}
+ Rewrite & 35.49& \textbf{29.19}& 4.66 & 36.35& 24.99& 4.93 & 36.21& 26.61&4.76\\
\midrule
\multicolumn{10}{l}{\textbf{Multi-Modal}}\\
 CombineMM& 37.65& 26.79& 3.89 & 38.61& 27.22& 3.92 & 37.12& \textbf{30.38}&3.79\\
 \rowcolor{mintgreen}
 + Rewrite & \textbf{41.09}& 25.32& 4.71 & 37.57& 25.92& 4.90 & \textbf{38.56}& 29.11&4.79\\
 \bottomrule
 \multicolumn{10}{l}{\scriptsize{*Notes: R-L denotes Rouge-L F1; RQG denotes RQUGE. Higher means better for all metrics.}}
\end{tabular}
\caption{Results on DL-Intro.}
\label{tab:results_dl}
\end{table}

\subsection{Main Results}
Results on LLM-Frontier and DL-Intro are shown in Table~\ref{tab:results_llm} and Table~\ref{tab:results_dl}, respectively. 

\paragraph{Using all transcripts or keyframes as context is suboptimal for QG.} In most cases (except for GPT on LLM-Frontier), using all transcripts or keyframes as context leads to the lowest performance, indicating that excessive or unfiltered context negatively impacts question quality. While the best fixed-length context window (Rule-Best) outperforms full context, our LLM-based CoT selection often yields better context. Specifically, direct extraction (Direct), as a basic strategy, outperforms Rule-Best with all models on LLM-Frontier, but underperforms Rule-Best on DL-Intro. However, CoT consistently improves over Direct, yielding higher NLI scores than All and Rule-Best.
These results show the effectiveness of CoT in dynamically selecting relevant context for QG.

\paragraph{Rewriting context improves relevancy and answerability.}
Rewriting improves NLI scores in most (24 out of 30) settings. For instance, CoT with rewriting reaches 38.36, up from 32.78 without rewriting, using GPT on DL-Intro. A similar trend is observed in keyframe-based context. The gains suggest that rewriting helps better align questions with instructional goals. Notably, rewriting consistently improves on RQUGE score. This suggests LLM context rewriting, as the essential component of our framework, is effective in re-organizing extracted transcripts or keyframes into answer-containing knowledge points that serve as concise and relevant context.

\paragraph{Transcripts are more useful context than keyframes, moreover, combining both modalities further improves question quality.}
Using transcripts alone consistently outperforms keyframe-based contexts across all settings. 
Moreover, the best performance is achieved when combining both modalities. On DL-Intro, CombineMM reaches an NLI of 37.65 without rewriting and 41.09 with rewriting, marking the highest scores overall. This demonstrates the complementary nature of the lecturer's speech and slides and the effectiveness of the rewriting step in enhancing question quality.

\subsection{Impacts of Rule-Based Context Window}

\begin{figure}[ht]
    \centering
    \begin{subfigure}{0.4\linewidth}
        \centering
        \includegraphics[width=\linewidth]{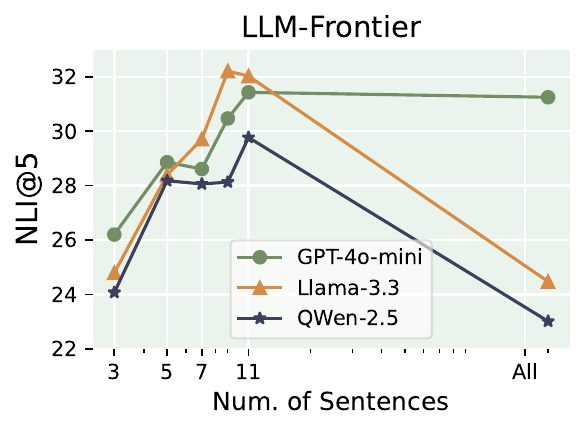}
    \end{subfigure}
    \vspace{0.5em} 
    \begin{subfigure}{0.4\linewidth}
        \centering
        \includegraphics[width=\linewidth]{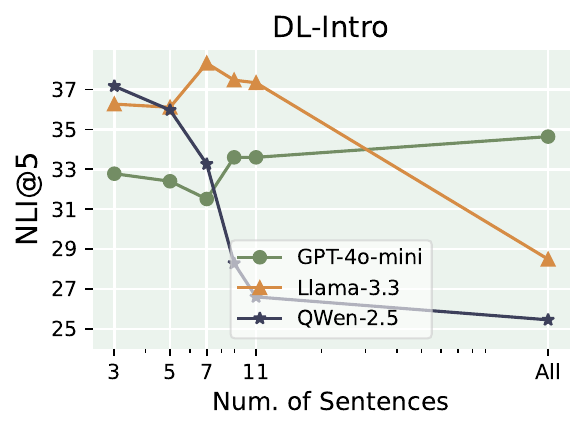}
    \end{subfigure}
    \caption{Increasing context window does not always improve question generation.}
    \label{fig:context_window}
\end{figure}

As demonstrated in Figure~\ref{fig:context_window}, varying context window sizes have a notable influence on the quality of generated questions, and simply expanding the window often fails to yield consistent improvements. Moreover, providing the entire context in most cases did not lead to the highest performance, suggesting that using all available information can introduce noise or dilute key details for question formation. We also observed that the optimal context window size differed across both datasets and models, indicating that there is no one-size-fits-all approach. These findings highlight the need for adaptive and selective context retrieval strategies, rather than relying on a static, predetermined context window.

\subsection{Case Study}

Table~\ref{tab:case_study} shows a test case from LLM-Frontier. In this example, based on the target answer, the instructor intended to assess the operation of ReLU activation function. With All Transcript as context, the generated question involves a general concept ``nonlinearity'' and an irrelevant concept ``convolutional neural network''. The CoT context selection has correctly identified the relevant concept of ReLU, but still uses inaccurate wording ``negative values'' in the generated question. The rewriting step and the integration of both modalities have rectified this issue, further aligning the generated question with instructor's intent. The complete example is provided in Table~\ref{tab:case_study_complete}.

\begin{table}[ht]
\centering
\scriptsize
\begin{tabular}{p{5.3in}}
\toprule
\textbf{Reference Question}: How does the ReLU activation function work? \\
\textbf{Target Answer}: It transforms all negative values into 0, while leaving positive values intact. \\
\midrule
\textbf{Generated Questions with Different Context Settings} \\ 

\textbf{All Transcript:} 
What does the \colorbox{lightyellow}{nonlinearity operation} in a \colorbox{lightyellow}{convolutional neural network} do to \colorbox{lightpink}{negative values}? (NLI: 15.88) \\

\textbf{CoT:} 
What does the \colorbox{mintgreen}{\underline{rectified linear unit (ReLU) activation function}} do to \colorbox{lightpink}{negative values} in the output? (NLI: 41.32) \\

\textbf{CoT+Rewrite:} 
What is the primary transformation performed by the \colorbox{mintgreen}{\underline{ReLU function}}? (NLI: 42.77) \\

\textbf{CombineMM:} 
In the context of deep learning, what is the primary role of the \colorbox{mintgreen}{\underline{ReLU activation function}}? (NLI: 40.32) \\

\textbf{CombineMM+ Rewrite:} 
How does \colorbox{mintgreen}{\underline{ReLU}} affect \colorbox{mintgreen}{\underline{values}} during its operation? (NLI: 47.95) \\ \midrule

\multicolumn{1}{l}{\textbf{Contexts from CombineMM}} \\ 

\multicolumn{1}{p{5.3in}}{ \textbf{Extracted}: \textbf{\textit{Contextual summary}}: In the context of convolutional neural networks, activation functions play a critical role ...  \textbf{\textit{From Transcript Excerpt}}: ``a common activation for convolutional neural networks are, rectified linear units or relu, right? think of this as really just deactivating pixels in your feature map that are negative, right? so \colorbox{mintgreen}{\underline{anything that's positive gets passed on and anything that's negative gets set to zero.}}''
\textbf{\textit{From Keyframe Excerpt}}: ``... Here's a summary of the key points:  1. **Application**: ...  4. **Graph Representation**: The function \( g(z) = \max(0, z) \) illustrates how ReLU operates, where values less than zero are set to zero. ...''} \\

\multicolumn{1}{p{5.4in}}{\textbf{Rewritten}: 1. Activation functions are crucial in convolutional neural networks as they introduce nonlinearity into the model.
2. The rectified linear unit (ReLU) is a common activation function that is applied after every convolution operation in CNNs.
3. \colorbox{mintgreen}{\underline{ReLU operates}} as a pixel-by-pixel transformation, where ``\colorbox{mintgreen}{\underline{It transforms all negative values into 0, while leaving positive values intact.}}''}\\

\bottomrule
\end{tabular}
\caption{An example in LLM-Frontier. With extracted and rewritten contexts, generated questions contains more \colorbox{mintgreen}{\underline{relevant information}} aligned with instructor's intent and less \colorbox{lightpink}{inaccurate} or \colorbox{lightyellow}{irrelevant} information.}
\label{tab:case_study}
\end{table}

\section{Conclusion}

In this work, we address the challenge of constructing appropriate contexts for generating educationally aligned questions. First, we construct a dataset of instructor-written quiz questions based on real-world classroom lectures, highlighting the limitations of existing EQG solutions when applied in realistic settings with long and noisy contexts of lecture speech. To bridge this gap, we propose COSER, an LLM-based EQG framework which first extracts relevant context pieces from full lecture content, then rewrites them into answer-containing atomic statements, and finally integrates multi-modal information from both lecture transcripts and video frames of slides for QG. Experiments with three LLMs demonstrate that COSER produces more relevant and concise contexts, thereby improving the relevancy and educational alignment of the generated questions.

\section*{Ethics Statement}
All lecture speakers provided informed consent for their recorded content. The study protocol was reviewed and approved by our Institutional Review Board (IRB). We carefully screened the collected video data to ensure that no student’s identifiable features (e.g., faces or other biometric details) remained in the dataset. All data handling procedures were conducted in compliance with relevant privacy regulations and institutional guidelines to protect participant confidentiality.

\section*{Limitation}

\paragraph{The size of our dataset.} As we aim to collect high-quality instructor-created quiz questions, the data collection is time-consuming, resulting a limited amount of data instances (about 500 questions) in our dataset AIRC. To expand AIRC, we are continuing the process of annotating more questions and plan to release the dataset for future use in the final version of the paper.

\paragraph{The difficulty of answerability-based evaluation.} Answerability-based metrics, such as RQUGE, use the question answering (QA) performance as a proxy to evaluate the quality of generated questions. However, the reliability of these methods depends on the model's QA capability, which is challenging in long-context setting. RQUGE has a context length limit of 512 tokens, thus we had to cut off exceeding contexts and might lose relevant details. Future work could consider using more advanced long-context QA models for evaluating answerability.

\paragraph{The integration of multi-modalities.} As a pilot study of video-based answer and timestamp-aware question generation, we explored the combination of audio and video by converting them into textual modality. This might lose rich information hidden in their original modalities. In future work, we will explore the use of Vision Language Models in processing transcript and keyframes as two modalities.

\bibliography{custom}
\bibliographystyle{colm2025_conference}

\appendix
\section{Appendix}

\subsection{Detailed Results on Question Pair Scoring}
Table~\ref{tab:better_eval_metric} presents the results of identifying paraphrasing question pairs using different evaluation metrics or models.
\begin{table}[ht]
\centering

\begin{tabular}{ p{4cm} | r  r | r  r }
\toprule
 & \multicolumn{2}{c}{\textbf{MEQ}}& \multicolumn{2}{c}{\textbf{QQP}}\\

\textbf{Metric / Model} & \textbf{F1} & \textbf{Acc.} & \textbf{F1} & \textbf{Acc.} \\
\midrule
\textbf{sileod/deberta-v3-large-tasksource-nli} & 97.71& 97.67& 86.91& 87.00\\

\textbf{sileod/deberta-v3-base-tasksource-nli} & 97.51& 97.50& 85.14& 85.33\\

\textbf{sileod/deberta-v3-small-tasksource-nli} & 95.65& 95.67& 77.57& 76.00\\

\textbf{MoritzLaurer/mDeBERTa-v3-base-xnli-multilingual-nli-2mil7} & 92.10& 92.17& 75.86& 75.50\\

\textbf{GPT3.5-0shot} & 88.89& 87.50& 81.82& 81.33\\

\textbf{GPT3.5-8shot} & 96.93& 96.83& 54.16& 67.83\\

\textbf{ParaScore} & 66.67& 50.00& 75.93& 72.00\\

\textbf{ROUGE-L} & 66.67& 50.00& 76.16& 72.67\\

\textbf{BLEU} & 22.19& 13.50& 53.91& 62.67\\

\textbf{BLEU-1} & 66.67& 50.00& 75.38& 70.50\\

\textbf{BLEU-2} & 59.98& 42.83& 77.37& 73.00\\

\textbf{BERTScore} & 66.67& 50.00& 78.79& 74.33\\
\bottomrule
\end{tabular}

\caption{Performance of exising evaluation metrics in accurately identifying pairs of paraphrased questions.}
\label{tab:better_eval_metric}
\end{table}

\subsection{Evaluation Details of RQUGE}
\paragraph{RQUGE} RQUGE \citep{mohammadshahi-etal-2023-rquge} is a reference-free metric based on answerability. Given a context, gold
answer span, and the candidate question, RQUGE computes an acceptance score scaled from 1 to 5. Due to the context length limitation of the RQUGE model, we truncate input contexts that exceed 512 tokens. For the transcript setting, we extract a context window centered around the timestamp, with the timestamp positioned at 75\% of the window. For extracted and rewritten context settings, we truncate content from the end of the text. 

\subsection{Prompt Templates}\label{sec:qg_prompt}

\begin{promptbox}{Prompt Template for Direct Context Selection from Transcripts}

You are tasked with extracting contexts from the given lecture transcript for generating a quiz question. You'll be provided the following information:\\

- Lecture Transcript: A lengthy text which is the transcript of a lecture.\\
- Answer: A word, phrase, or sentence that serves as the answer to a potential quiz question.\\
- Timestamp: A sentence ID to indicate the time associated with the quiz question.\\

Your output must include the following components:\\

1. Contextual Summary:\\
- Provide a brief and clear summary of the main ideas or topics discussed immediately before the selected transcript excerpt.\\
- The summary should frame and support the context for the chosen excerpt, enabling a quiz creator to understand the background necessary for formulating the question.\\

2. Extracted Transcript Excerpts:\\
- Select one or more segments directly from the lecture transcript.\\
- It must be self-contained, clearly providing all information required for ** creating ** a quiz question with the provided the answer ``\{answer\}''.\\
- Might be near the timestamp sentence (ID \{context\_id\}).\\
- Each segment should contain a ** complete ** piece of information or atomic knowledge, including the subjects that pronouns refer to.\\
- Is concise and only includes the necessary details relevant to the quiz question, spanning approximately 3 to 6 sentences.\\
- Should match the original transcript text exactly. Do not alter the text in any way.\\

Lecture Transcript:
\{context\}

Answer to quiz question:
\{answer\}

Timestamp: ID \{context\_id\}\\

Output format:

``Extracted Context: [extracted transcript excerpt here]''

\end{promptbox}

\begin{promptbox}{Prompt Template for Chain-of-Thought Context Selection from Transcripts}
... (Same instruction as Direct extraction)\\

Let's get started! Please think step by step. First, list all the transcripts that are relevant to the answer and timestamp. Second, double check and adjust each context on a sentence-by-sentence basis based on the requirements. You can add or remove sentences. Lastly, list the final extracted context.\\

Output format:

``Reasoning: [your reasoning process] 
Extracted Context:
Contextual summary: [contextual summary here]
From Transcript Excerpt: [first final extracted context] [second final extracted context, if applicable]''

\end{promptbox}

\begin{promptbox}{Prompt Template for Direct Context Selection from Keyframes}
You are tasked with extracting contexts from the given lecture materials for generating a quiz question. You'll be provided the following information:

- Slide Description: Description of a set of slides of the lecture.\\
- Answer: A word, phrase, or sentence that serves as the answer to a potential quiz question.\\
- Timestamp: A slide ID to indicate the time associated with the quiz question.\\

Output the slide description that meets the following requirements:\\
- Is of the slides that are most relevant to the quiz question, where the answer would be ``\{answer\}''.\\
- Is sufficient and provides all relevant contexts for ** creating ** a quiz question.\\
- Might be near the slide ID \{keyframe\_id\}.\\
- Should contain the descriptions of 1 to 3 slides.\\
- Should match the original slide description text exactly. Do not alter the text in any way.\\

Slide Description:
\{context\}

Answer to quiz question:
\{answer\}

Timestamp: ID \{keyframe\_id\}\\

Output format:

``Extracted Context: From Slides: [extracted slide excerpt]''

\end{promptbox}

\begin{promptbox}{Prompt Template for Chain-of-Thought Context Selection from Keyframes}
... (Same instruction as Direct extraction)\\

Let's get started! Please think step by step. First, list all the keyframes that are relevant to the answer and timestamp. Second, double check and adjust each context on a sentence-by-sentence basis based on the requirements. You can add or remove sentences. Lastly, list the final extracted context.\\

Output format:

``Reasoning: [your reasoning process]  
Extracted Context: From Slides: [final extracted slide excerpt]''

\end{promptbox}

\begin{promptbox}{Prompt Template for Context Rewriting}
You task is to rewrite the contexts extracted from a lecture for creating quiz questions assessing the understanding of the lecture. You'll be provided the following information:

- Extracted Context: A piece of context extracted from the lecture transcript or slides.
- Answer: A word, phrase, or sentence that serves as the answer to a potential quiz question.\\

You can rewrite the transcripts into 3-5 statements. Each statement should meet the following requirements:\\
- Most importantly, the provided answer ``{answer}'' must appear WORD-FOR-WORD.\\
- Should be a piece of knowledge which could be an important learning point to be assessed in a quiz.\\
- Should preserve the original lecture language. Keep all technical/domain-specific terms exactly as presented. Try NOT to alter the text. Only allows changes for: Grammatical connections, removing reference words (this, that, it, etc.), and converting indirect to direct speech.\\
\\
Notes:\\
- If no statements can be directly rewritten from the transcripts with the provided answer, then you may write a statement inferred from the context, as long as it contains the provided answer WORD-FOR-WORD.\\
- The final statements should be rewritten from the transcripts, but you can integrate the information from the keyframes and contextual summary into it.\\
- You don't have to incorporate all the information into rewritten statements. Just focus on the ones that are relevant to the provided answer.\\
- Importantly, the goal is to capture a structured set of knowledge points of different levels of granularity, ranging from general or conceptual statements to specific or technical details.\\

Contexts:
\{context\} \\
Answer: \{answer\}\\

Please think step by step. Please respond in this format:\\

``Reasoning: [your reasoning process]  
Rewritten Contexts: [list of final rewritten statements here]''
    
\end{promptbox}

\begin{promptbox}{Prompt Template for Question Generation}
    You're an experienced STEM teacher. Your task is to generate multiple-choice quiz questions from the lecture content. All questions should naturally lead to the provided correct answer. The questions should be concise and preserve the original language of the lecture content, and should be interrogative sentences. You don't have to incorporate all the information from the lecture content into the questions. You can write 5 questions in total.\\
    
    Use this format for each question:\\
    ``Q1. [Question text] \\
    A) [Option A] \\
    B) [Option B] \\
    C) [Option C] \\
    D) [Option D]'' \\

    Lecture Content: \{context\}\\
    Correct Answer: \{answer\} \\
    Timestamp: \{context\_id\} (This line is only provided in All transcripts or keyframes setting) \\

    Please provide your questions below.
    
\end{promptbox}

\subsection{Video Keyframe Detection Algorithm}\label{sec:keyframe_detection_algo}
We use Algorithm \ref{alg:keyframe_detection} to detect and extract keyframes from lecture videos.
\begin{algorithm}
\caption{Keyframe Extraction}
\begin{algorithmic}[1]
\Procedure{ExtractKeyFrames}{video, outputDir, fps}
    \State Initialize $frameCount \gets 0$, $keyframeIndex \gets 0$
    \State $prevHash \gets$ ArbitraryInitialHash
    \State Create $outputDir$ if it does not exist
    \State Open $video$ as $cap$
    \State $success, frame \gets$ ReadFirstFrame($cap$)
    \While{$success$}
        \For{$i = 1$ to $N$} \Comment{Skip $N$ frames}
            \State $success, frame \gets$ ReadNextFrame($cap$)
        \EndFor
        \If{not $success$}
            \State \textbf{break}
        \EndIf

        \State $currentHash \gets$ PerceptualHash($frame$)

        \If{$keyframeIndex = 0$ \textbf{or} Distance($currentHash$, $prevHash$) $> \delta$}
            \State $timestamp \gets \lfloor frameCount / fps \rfloor$
            \State Save $frame$ as image in $outputDir$ with name based on $timestamp$
            \State $keyframeIndex \gets keyframeIndex + 1$
        \EndIf
        \State $prevHash \gets currentHash$
        \State $frameCount \gets frameCount + 1$
    \EndWhile
\EndProcedure

\Function{Distance}{$h_1, h_2$}
    \If{Length($h_1$) $\neq$ Length($h_2$)}
        \State \textbf{raise} Error
    \EndIf
    \State \Return $\sum_{i} | h_1[i] - h_2[i] |$
\EndFunction
\end{algorithmic}
\label{alg:keyframe_detection}
\end{algorithm}

\subsection{A Complete Example of Case Study}
\begin{table}[ht]
\centering
\small
\renewcommand{\arraystretch}{1.5} 
\begin{tabular}{@{}p{0.8in} p{4in}@{}}
\toprule
\multicolumn{2}{l}{\textbf{Generated Questions with Different Context Setting}} \\ \midrule

\textbf{All Transcript} & 
What does the nonlinearity operation in a convolutional neural network do to negative values? (NLI: 0.1588) \\

\textbf{CoT} & 
What does the rectified linear unit (ReLU) activation function do to negative values in the output? (NLI: 0.4132) \\

\textbf{CoT + Rewrite} & 
What is the primary transformation performed by the ReLU function? (NLI: 0.4277) \\

\textbf{CombineMM} & 
In the context of deep learning, what is the primary role of the ReLU activation function? (NLI: 0.4032) \\

\textbf{CombineMM + Rewrite} & 
How does ReLU affect values during its operation? (NLI: 0.4795) \\ \midrule

\multicolumn{2}{l}{\textbf{Contexts by CoT}} \\ 

\multicolumn{2}{p{5in}}{\textbf{Extracted:} \textit{From transcript}: ``so anything that's positive gets passed on and anything that's negative gets set to zero. so think of it almost like a thresholding function.'' \textit{Contextual summary}: In this part of the lecture, the speaker discusses the application of a nonlinearity in convolutional neural networks, specifically focusing on the rectified linear unit (ReLU) activation function. This function is essential for introducing nonlinearity into the model, allowing it to handle complex data patterns. The speaker emphasizes how the ReLU function operates by deactivating negative pixel values in the feature map, effectively transforming them to zero while preserving positive values.} \\

\multicolumn{2}{p{5in}}{\textbf{Rewritten: }
1. The rectified linear unit (ReLU) activation function is crucial for introducing nonlinearity into convolutional neural networks. 
2. The ReLU function operates by deactivating negative pixel values in the feature map.
3. Specifically, the ReLU function applies a thresholding effect, where `It transforms all negative values into 0, while leaving positive values intact.'
4. This transformation allows the model to effectively handle complex data patterns by preserving only the positive values.} \\ \midrule

\multicolumn{2}{l}{\textbf{Contexts by CombineMM}}\\

\multicolumn{2}{p{5in}}{Extracted: Contextual summary: In the context of convolutional neural networks, activation functions play a critical role in introducing nonlinearity into the model. One common activation function used is the rectified linear unit (ReLU), which is essential for allowing the network to learn complex patterns in the data. The ReLU function specifically modifies the output of neurons by setting all negative values to zero, which helps in maintaining the positive values for further processing.  From Transcript Excerpt: ``a common activation for convolutional neural networks are, rectified linear units or relu, right? think of this as really just deactivating pixels in your feature map that are negative, right? so anything that's positive gets passed on and anything that's negative gets set to zero.''
From Keyframe Excerpt:``The image appears to discuss the concept of nonlinearity in deep learning, specifically focusing on the Rectified Linear Unit (ReLU) activation function. Here's a summary of the key points:  1. **Application**: ReLU is applied after every convolution operation in convolutional neural networks (CNNs).  2. **Functionality**: It is a pixelbypixel operation that replaces negative values with zero, making it a nonlinear activation function.  3. **Visual Representation**: The image shows an input feature map where:     Black represents negative values.     White represents positive values.     The rectified feature map shows only nonnegative values.  4. **Graph Representation**: The function $ g(z) = \max(0, z) $ illustrates how ReLU operates, where values less than zero are set to zero.  5. **Implementation**: The last part of the image includes a reference to the Keras library for implementing the ReLU layer.  This serves to highlight the importance of adding nonlinearities to neural networks to enhance their capacity to model complex relationships in data.} \\

\bottomrule
\end{tabular}
\caption{An Complete Example from LLM-Frontier.}
\label{tab:case_study_complete}
\end{table}

\end{document}